%% file: main.tex
\documentclass[10pt,twocolumn,letterpaper]{article}

\usepackage[pagenumbers]{cvpr}      
\usepackage{algorithm}
\usepackage{algorithmic}
\usepackage{newfloat}
\usepackage{multirow}
\usepackage{listings}
\usepackage{amsmath,amssymb}
\usepackage{booktabs}
\usepackage{colortbl}
\usepackage{subcaption}

\definecolor{cvprblue}{rgb}{0.21,0.49,0.74}
\usepackage[pagebackref,breaklinks,colorlinks,allcolors=cvprblue]{hyperref}

\def\paperID{15697} 
\def\confName{CVPR}
\def\confYear{2025}

\title{LaPIG: Cross-Modal Generation of Paired Thermal and Visible Facial Images}

\author{Leyang Wang\\
University College London\\
London, UK\\
{\tt\small leyang.wang.24@ucl.ac.uk}
\and
Joice Lin\\
Xiamen University\\
Xiamen, China\\
{\tt\small 23020201153777@stu.xmu.edu.cn}
}

\begin{document}
\maketitle
\input{sec/0_abstract}    
\input{sec/1_intro}
\input{sec/2_relatework}
\input{sec/3_method}
\input{sec/4_experiments}
\input{sec/5_conclusion}
{
    \small
    \bibliographystyle{ieeenat_fullname}
    \bibliography{main}
}


\end{document}

%% file: sec/0_abstract.tex
\begin{abstract}
The success of modern machine learning, particularly in facial translation networks, is highly dependent on the availability of high-quality, paired, large-scale datasets. However, acquiring sufficient data is often challenging and costly. Inspired by the recent success of diffusion models in high-quality image synthesis and advancements in Large Language Models (LLMs), we propose a novel framework called LLM-assisted Paired Image Generation (LaPIG). This framework enables the construction of comprehensive, high-quality paired visible and thermal images using captions generated by LLMs. Our method encompasses three parts: visible image synthesis with ArcFace embedding, thermal image translation using Latent Diffusion Models (LDMs), and caption generation with LLMs. Our approach not only generates multi-view paired visible and thermal images to increase data diversity but also produces high-quality paired data while maintaining their identity information. We evaluate our method on public datasets by comparing it with existing methods, demonstrating the superiority of LaPIG.
\end{abstract}

%% file: sec/1_intro.tex
\section{Introduction}
\label{sec:intro}

Modern generative models have made significant contributions to various fields, including image synthesis \cite{rombach2022high,gu2022vector,meng2021sdedit}, video generation \cite{ho2022video,singer2022make,khachatryan2023text2video}, speech synthesis \cite{tan2024naturalspeech,li2019neural}, and text generation \cite{achiam2023gpt,touvron2023llama}. Among these areas, conditional image generation is especially notable due to its wide range of applications, such as the transformation between visible and thermal images \cite{nair2023t2v,yang2021infrared,shamsolmoali2021image,li2024deep}.

The effectiveness of these generative models is largely attributed to the availability of high-quality, large-scale paired datasets \cite{changpinyo2021cc12m,schuhmann2022laion}. However, existing public datasets are limited for various training tasks. Simple methods for creating these datasets require extensive time and financial cost. For instance, collecting visible images involves recruiting volunteers, obtaining consent, setting up equipment, and adhering to privacy standards. Collecting thermal images adds complexities such as managing machine noise and varying environmental conditions, which can affect data accuracy. Given these challenges and advances in generative techniques, an important question emerges: \textit{Can we generate such paired images within the framework of modern generative models?}

\begin{figure}[t!]
\centering
\includegraphics[width=0.45\textwidth]{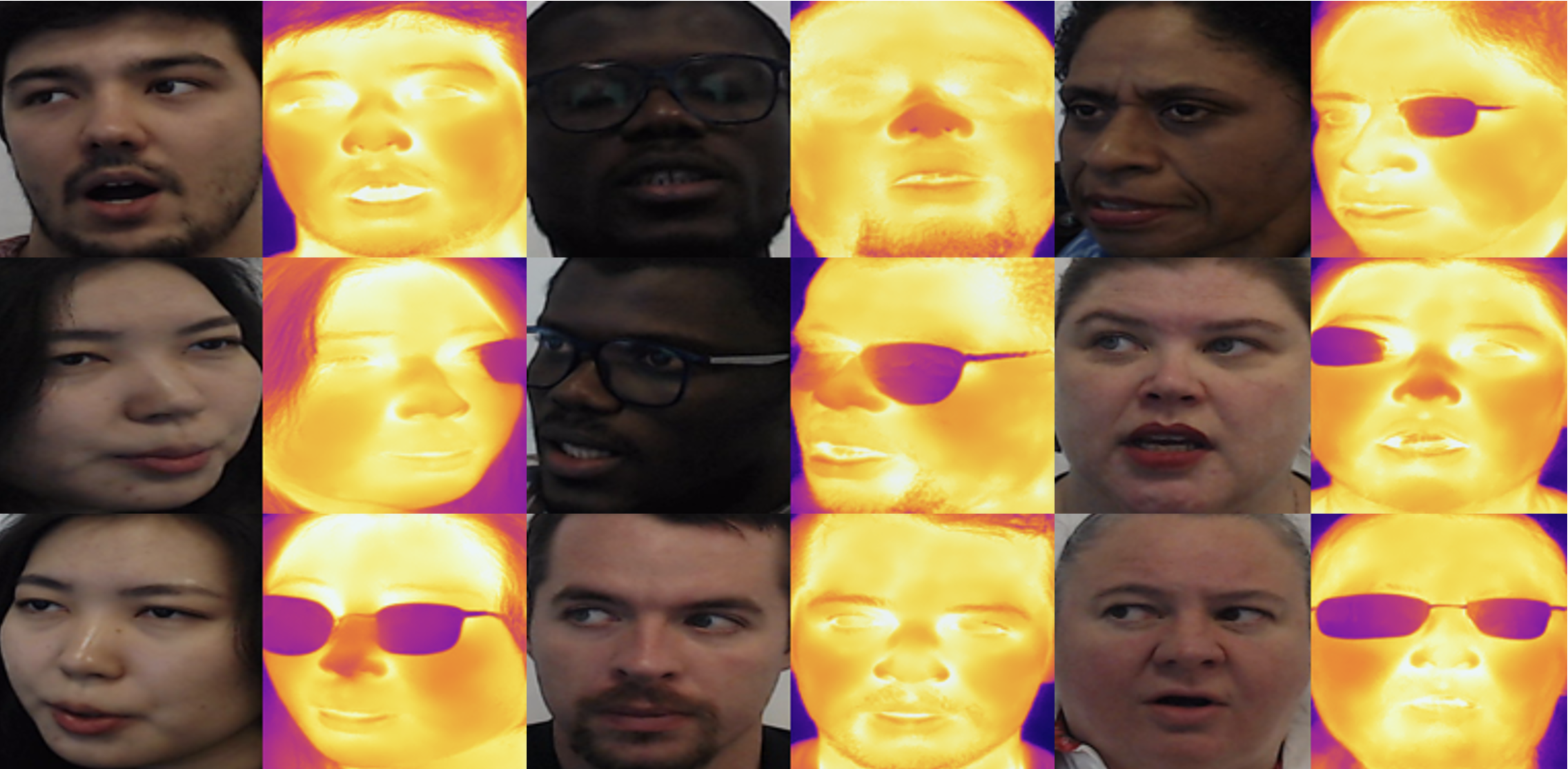}
\caption{{\bf Cases from previous image translation methods. } The image above shows the visible-to-thermal transformation effects of pix2pix \cite{isola2017image} and T2V-DDPM \cite{nair2023t2v}. It is evident that artifacts often appear in the generated data, while the areas that should have produced features such as glasses or beards are ineffective.
}
\label{fig:badcase}
\end{figure}

Traditional methods for identity-preserving generation use Generative Adversarial Networks (GANs), such as Style-GAN\cite{karras2019style,karras2020analyzing,karras2021alias}. These methods involve learning a decoder controlled by low-dimensional latent codes to generate images. However, a persistent challenge with StyleGAN is identity drift during latent space manipulations. GAN-based methods for converting visible images to thermal images often fail due to the adversarial nature of their training strategy\cite{saxena2021generative,gui2021review,kammoun2022generative}. The current state-of-the-art models for identity-preserving generation fall into two main categories. The first category includes Dreambooth-based\cite{ruiz2023dreambooth} methods, which require fine-tuning model parameters with dozens of images of the same identity (ID). Customizing each ID in this way takes long time and demands substantial computing resources, especially as model sizes increase. The second category relies on trained visual encoders or hypernetworks that encode input ID images as embeddings or LoRA\cite{hu2021lora} weights within the model, eliminating the need for additional training for generation. Notable methods in this category include PhotoMaker\cite{li2024photomaker} and Arc2face\cite{papantoniou2024arc2face}. 

Furthermore, to enable transformations across different modalities, current methods such as GAN-based facial translation networks\cite{isola2017image} and diffusion-based networks\cite{nair2023t2v} have been explored. However, as illustrated in Figure \ref{fig:badcase}, existing image translation methods often memorize dataset-specific features that are irrelevant to the target domain and lack sensitivity to fine details, resulting in suboptimal thermal image translations. The introduction of Latent Diffusion Models(LDMs)\cite{rombach2022high} offers a promising alternative. By conditioning on text, images, and other inputs, LDMs can generate high-quality visuals, presenting a potential solution to these limitations.


In light of these developments and existing challenges, we propose a framework named LLM-assisted Paired Image Generation(LaPIG) to synthesize paired images in different modalities with LLM-assisted. Inspired by PhotoMaker\cite{li2024photomaker} and Arc2face\cite{papantoniou2024arc2face}, we propose a training-free method for generating identity-preserved photos using ID embeddings produced by ArcFace backbone as first part of LaPIG. We enhance the quality of synthesis by incorporating captions generated by an LLM to refine control over identity preservation by enriching textual descriptions with a comprehensive identity-specific information. The resulting visible images are then processed by an LDM-based visible to thermal image(V2T) translation network to produce the corresponding thermal images as the second part of proposed system. 

Our method has demonstrated its effectiveness in both quality and time efficiency. Compared to manual data collection, LaPIG significantly reduces time and financial costs. Furthermore, in comparison to traditional GAN-based methods, LaPIG generates higher-quality images. The synthesized paired data have multiple usages in real-world, for example, some downstream application like facial recognition.


In summary, our contributions are as follows:
\begin{itemize}
    \item We introduce LaPIG, a novel framework with LLM-assisted that enable generating multi-view visible images from existing ID images while preserving identity details and creating matching thermal images.
    \item We develop an innovative method for generating visible images while maintaining identity information by incorporating long-CLIP encoder and LLMs.
    \item To the best of our knowledge, LaPIG is the first to apply LDM for the task of translating visible face images to thermal face images with identity preservation.
    \item We rigorously evaluate LaPIG through extensive experiments, demonstrating its effectiveness and superiority over existing methods.
\end{itemize}

%% file: sec/2_relatework.tex
\section{Related Works}
\label{sec:formatting}

In this section, we discuss existing work related to our research. We will delve into text-to-image generation, followed by a discussion of GAN-based methods and identity-preserving image generation.
\subsection{Text Conditioned Image Generation}
Recently, Diffusion models \cite{ho2020denoising,song2020denoising} have demonstrated state-of-the-art performance in text-to-image generation \cite{rombach2022high,zhang2023adding, kawar2023imagic,yang2024mastering} due to their ease of training with simple and efficient loss functions and their ability to generate highly realistic images. DALLE-2 \cite{ramesh2022hierarchical}, for example, employs a diffusion model conditioned on image embeddings, allowing text-based and image-based prompts for image generation. In terms of controllable generation, ControlNet \cite{zhang2023adding} adds spatial conditioning controls to pre-trained text-to-image DMs using zero-initialized convolution layers. Furthermore, Versatile Diffusion \cite{xu2023versatile} introduces a unified multi-flow diffusion approach, capable of supporting text-to-image, image-to-text, and various other tasks within a single model.


\subsection{GAN-based Face Translation Networks.}
Generative Adversarial Networks (GANs) \cite{goodfellow2014generative, isola2017image, radford2015unsupervised, esser2021taming} are a type of generative model that concurrently train a discriminator to determine whether images are real or fake and a generator to reduce this discernment loss. SAGAN \cite{di2019sagan} uses self-attention to create visible faces from thermal images for identity matching, highlighting the unique identity cues in thermal images. Axial-GAN \cite{immidisetti2021axial-gan} applies axial-attention to capture long-range dependencies with transformer technology, allowing for the creation of detailed visible images. VPGAN \cite{mei2022VPGAN} relies on known facial features from the visible spectrum to streamline the learning process. UVCGAN \cite{torbunov2023uvcgan} boosts image translation quality and variety by combining ViT's non-local pattern recognition with Cycle-GAN's consistency rules. HiFaceGAN \cite{yang2020hifacegan} progressively refines facial details guided by hierarchical semantics, tackling the challenge of reconstructing faces with diverse degradation and backgrounds. However, training a GAN involves finding a Nash equilibrium in a game where regular gradients may not always effectively resolve the problem \cite{salimans2016improved}, which underscores the inherent challenges posed by their adversarial training strategy.

\begin{figure*}[t!]
\centering
\includegraphics[width=0.99\textwidth]{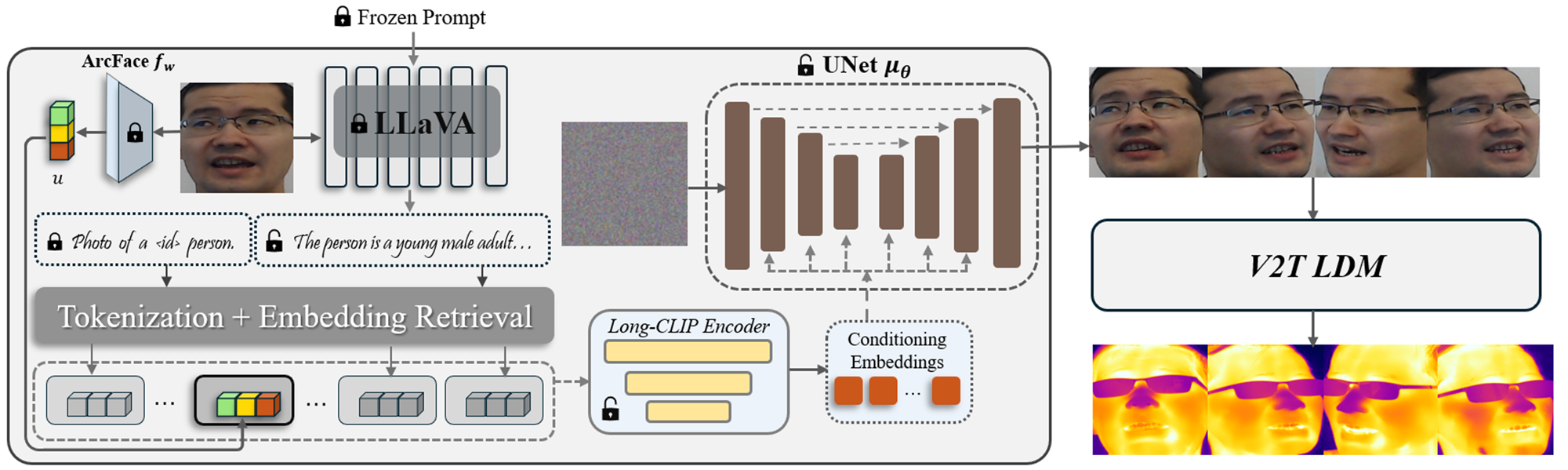}
\caption{{\bf Overview of LaPIG.} The LaPIG architecture we propose mainly consists of two parts: First is the proposed ID preserving synthesis model, which is used to generate a variety of facial images; second is V2T-LDM, responsible for converting the generated facial images into corresponding thermal images. To generate images that better match the original characteristics of the person, we used LLaVA to describe the person's age, gender, facial features, etc. }
\label{fig:pipeline}
\end{figure*}

\subsection{Identity Preserving Image Generation}
Identity-preserving generation aims to generate new images based on text descriptions while maintaining the original identity of the subject. This domain categorizes current methods into two primary types based on their need for fine-tuning during testing. The former methods involve fine-tuning pre-trained text-to-image models to integrate new elements from reference images\cite{gal2022image,kawar2023imagic,hu2021lora,ruiz2023dreambooth,chen2023disenbooth}, but this process is resource-intensive, time-consuming, and typically necessitates multiple subject images, thus limiting its practicality. The other type focuses on generating extensive domain-specific datasets and designing lightweight adapters to extract features from reference images\cite{li2024photomaker,wang2024instantid,Shi_2024_CVPR,wei2023elite,song2024imprint}. Notable implementations include Arc2Face\cite{papantoniou2024arc2face}, which uses the Arcface method \cite{deng2019arcface} to convert features from a human face into an ID vector, and then projects these using CLIP\cite{radford2021learning} for cross-modality control. Additionally, the IP-Adapter\cite{ye2023ip} utilizes a distinct cross-attention mechanism to separate text and image features, allowing the reference image to serve as a visual prompt. 

%% file: sec/3_method.tex
\section{Method}
In this section, we will discuss the entire paired images synthesis process using LaPIG step by step. First, we will introduce the preliminaries for the proposed method. Then, we will provide an overview of our pipeline, followed by a detailed analysis of each component. 


\subsection{Preliminaries}
\textbf{Latent Diffusion Models. }
DMs \cite{ho2020denoising, song2020denoising, song2020score} are structured in two primary phases: the diffusion phase and the denoising phase. The former introduces Gaussian noise into the data over various timesteps \( t \) following a pre-established schedule, while the latter involves a learnable model that reconstructs the data from the noisy distribution. The training of DMs involves a denoising autoencoder \( \epsilon_{\theta} \), which typically employs a UNet \cite{ronneberger2015u} as the backbone, is trained to minimize the loss function given by:
\begin{equation}\label{eq:diffloss}
    \mathcal{L}_{df} = \mathbb{E}_{x_t, t, \epsilon \sim \mathcal{N}(0, I)} \left[ \left\| \epsilon - \epsilon_{\theta}(x_t, \boldsymbol{c}, t) \right\|_2^2 \right]
\end{equation}
where \( x_t = \sqrt{\bar{\alpha}_t} x_0 + (1 - \bar{\alpha}_t) \epsilon \) represents the noisy data at timestep \( t \), \( \bar{\alpha}_t \) is a pre-defined function determining the extent of diffusion at \( t \), and \( \boldsymbol{c} \) represents conditioning variables such as text or images. The model enables the generation of images by progressively denoising from an initial random noise state.

Latent Diffusion Models (LDMs) \cite{rombach2022high}  leverage a Variational Autoencoder (VAE) \( \mathcal{E} \) to encode images into a reduced-dimensional latent space \( \mathcal{Z}=\{\mathcal{E}(x): x \in \mathcal{X}\} \), where \( \mathcal{X} \) denotes the space of possible images. The denoising autoencoder training \( \epsilon_{\theta} \) is then performed within this latent space, thereby increasing both efficiency and effectiveness in image synthesis. Importantly, LDMs incorporate a conditioning mechanism that utilizes cross-attention layers to project auxiliary features \( \boldsymbol{c} \) onto intermediate layers, thus facilitating more precise conditional noise prediction.
\vspace{0.1cm} \\
\textbf{Arcface }The utilization of pre-trained facial recognition networks to constrain face-related optimization or generation tasks has seen significant success in recent years. Additive Angular Margin Loss (ArcFace) \cite{deng2019arcface} is referred to as an angular margin. This margin-based softmax loss function is utilized in the training of face recognition systems. Let \(\theta_j\) represent the angle between the feature vector and the $j$-th classifier weight vector, ArcFace can be expressed as follows:
\begin{equation}\label{eq:arcface}
\mathcal{L}_{arc} = -\frac{1}{N} \sum_{i=1}^{N} \log \frac{e^{s(\cos(\theta_{y_i} + m))}}{e^{s(\cos(\theta_{y_i} + m))} + \sum_{j: j \neq y_i} e^{s \cos \theta_j}}
\end{equation}
where the variable \(y_i\) denotes the index of the ground truth label, \(m\) is the margin added to enhance feature discriminability, and \(s\) is a hyper-parameter that scales the input to the cosine function, thereby modulating the strictness of the classification boundary. 

Typically, ID similarity is employed as an auxiliary loss, often by extracting features from multiple layers of these networks. In this work, we aim to utilize these networks as frozen feature extractors in order to condition a generative model of facial images.


\subsection{Overview} 
\begin{algorithm}[!tbp]
    \caption{Inference Procedure of LaPIG.}
    \renewcommand{\algorithmicrequire}{\textbf{STEP1:}}
    \renewcommand{\algorithmicensure}{\textbf{STEP2:}}
    \begin{algorithmic}[1]
        \STATE \textbf{Prepare} Trained Synthesis Pipeline \textbf{Pipe}; ID Encoder $f_{w}$; V2T LDM pipeline $\textbf{LDM}$; ID Images $\mathcal{V}$; LLaVA $\mathcal{Q}$
        \REQUIRE Visible Images Synthesis
        \STATE Generate captions using LLaVA: $\boldsymbol{t} = \mathcal{Q}(\mathcal{V})$;
        \STATE Extract ID Vector: $\boldsymbol{u} = f_w(\mathcal{V})$;
        \FORALL{View Directions}
            \STATE Customize Prompt: $\textbf{p} = \text{view direction prompt}+\boldsymbol{t}$;
            \STATE Identity Preserving Generation: Append Set of Synthesized Visible Images $\hat{\mathcal{V}}$ with $\hat{v} = \textbf{Pipe}(\textbf{p},v,\boldsymbol{u})$;
        \ENDFOR
        \ENSURE V2T Translation
        \STATE Translate Synthesized Images from Visible Modality to Thermal Modality: $\hat{\mathcal{T}} = \textbf{LDM}(\hat{\mathcal{V}})$;
        \RETURN Paired data $\hat{\mathcal{V}}\times \hat{\mathcal{T}}$.
    \end{algorithmic}
    \label{alg:lapiginfer}
\end{algorithm}
Our goal is to synthesize multi-view paired images from existing photographs while preserving identity information to increase the diversity of data. Our approach can be divided into three main parts: {\it {(i)} identity-preserving visible image generation, {(ii)} thermal image translation, and {(iii)} caption generation using an LLM assistant.}

As described in Algorithm \ref{alg:lapiginfer}, we start with a set of ID images $\mathcal{V}$. We first feed the images to LLaVA to generate a comprehensive textual description of images identity. To create the first part of our paired images, we focus on identity-preserving image synthesis from \(v \in \mathcal{V}\). We start by obtaining an identity vector \(\boldsymbol{u} = f_w(v) \in \mathbb{R}^{512}\) using an Arcface network and add zero padding to \(\boldsymbol{u}\) to match the dimension of other tokens. The padded identity vector \(\hat{\boldsymbol{u}}\) is used as input with other tokens produced from text generated by LLaVA \cite{liu2023improvedllava,liu2023llava,liu2024llavanext} and mapped by the LongCLIP \cite{zhang2024long} encoder to the embedding space. This embedding, which includes identity and prompt information, guides the generator $\mu_{\theta}$ in synthesizing identity-preserved images.



Subsequently, we employ a trained LDM V2T translation network to translate thermal images \(x \in \mathcal{T}\) from synthesized visible images \(v \in \mathcal{V}\). We can obtain the paired data which contains images of multi-view direction as desired. Figure \ref{fig:pipeline} shows the overview of LaPIG. 

\subsection{Identity Preserving Synthesis with LLM-assisted}
Existing methods for identity-preserving synthesis have shown strong capabilities in generating identity-consistent images. A key feature of these methods is the use of ArcFace loss (\ref{eq:arcface}) to train an identity encoder \(f_w\), which can extract identity information as an identity vector \(\boldsymbol{u}\).

However, the goal is to synthesize multi-view facial images derived from existing photos to enhance data diversity. Prior studies have highlighted challenges with the CLIP encoder, such as misaligned signals and limitations on input text length. Recent findings suggest that the optimal prompt length for CLIP is under 20 tokens \cite{zhang2024long}, limiting its ability to handle complex textual prompts. These constraints, in turn, impede these models' performance when guided by lengthy textual inputs.

To address these challenges, we leverage LongCLIP \cite{zhang2024long}, a powerful plug-and-play alternative to CLIP for image generation. Additionally, recent advancements in large language models (LLMs), particularly multimodal models that combine a vision encoder with Vicuna for enhanced visual and language understanding, enable the automatic generation of textual prompts. Consequently, we use the Large Language and Vision Assistant(LLaVA)\cite{liu2023improvedllava,liu2023llava,liu2024llavanext} to generate long textual descriptions that incorporate identity-specific information.

In detail, as shown in Figure \ref{fig:pipeline}, our proposed method functions as follows: we use a pretrained ArcFace network \(f_{w}:\mathcal{V}\rightarrow\mathbb{R}^{512}\) as our ID encoder, where \(w\) is the weight, and employ the {\it stable-diffusion-v1-5} model as the UNet backbone \(\mu_{\theta}\) for visible image synthesis. For each visible image \(v\), we simultaneously input it into \(f_w\) to generate the identity vector and into LLaVA \(\mathcal{Q}\) to produce a comprehensive textual description. We then align dimensions to create an identity vector \(\hat{\boldsymbol{u}}\) and integrate it with the result tokens from \(\mathcal{Q}(v)\). Finally, the conditioning embedding produced by LongCLIP is used to guide the synthesis of the generator \(\mu_{\theta}\).

Our experiments demonstrate that LongCLIP can significantly enhance performance with comprehensive text prompts, generating higher-quality images than existing methods in our task and meeting our diversity requirements. Further details of the quantitative and qualitative analyses are available in the experiment section.

\subsection{V2T Latent Diffusion Models}
\begin{figure}[t!]
\centering
\includegraphics[width=0.45\textwidth]{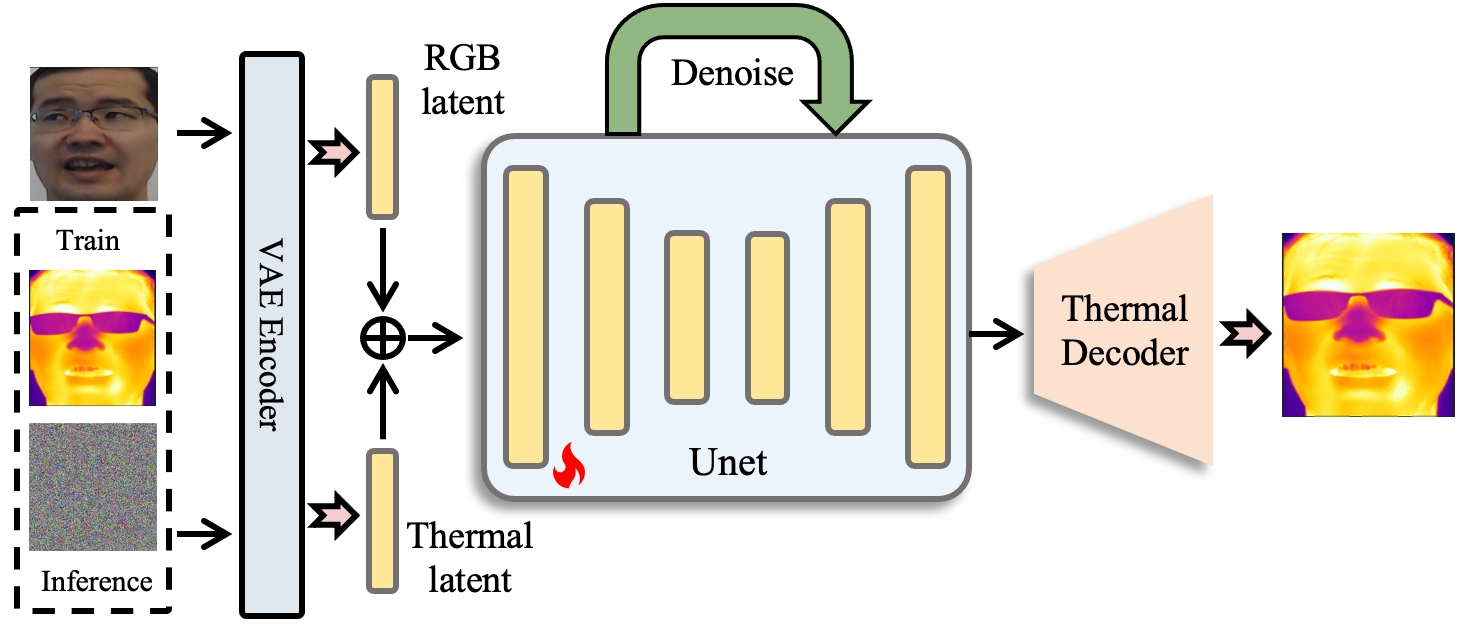}
\caption{{\bf Details of V2T LDM. }The process involves training and inference. In training, we combine the VAE thermal latent with the RGB image on the channel axis, sample at time t, and train Unet for noise prediction. In inference, we substitute the thermal image with Gaussian noise and denoise to get the predicted thermal image.}
\label{fig:ldm}
\end{figure}
DMs work as probabilistic models that approximate a data distribution \(p\) by gradually removing noise from a normally distributed variable, similar to reversing a fixed Markov Chain of length \(T\). However, regular DMs do not perform well in our visible-to-thermal (V2T) image translation task, as shown in the experiment part. In addition, DMs require a large amount of time for inference, which limits their usage.

Inspired by the success of DMs in thermal-to-visible image translation and the effectiveness of LDMs in high-quality image synthesis, we employ LDMs as the thermal-to-visible (T2V) translation network. The key advantage of LDMs is that they use a VQ-VAE for perceptual compression and diffusion models for semantic compression.

As detailed in Figure \ref{fig:ldm}, the V2T LDM process can be detailed as follows: within the available dataset \(D\), we first train a VQ-VAE consisting of an Encoder \(\mathcal{E}:\mathcal{X}\rightarrow\mathcal{Z}\) and a Decoder \(\mathcal{D}:\mathcal{Z}\rightarrow\mathcal{X}\). Here, \(\mathcal{Z} = \{\mathcal{E}(x) : x\in\mathcal{X}\}\) represents the latent space. The training process involves maximizing evidence lower bound (ELBO) and aligning the embedding vectors closely with the encoder output.  Subsequently, we map both the paired data \(v\) and \(x\) to the latent space using \(\mathcal{E}\), and then train a diffusion model $\boldsymbol{\epsilon}_{\boldsymbol{\vartheta}}$, typically parameterized with a UNet, to learn the conditional distribution \(p_{\boldsymbol{\vartheta}}(\mathcal{E}(x_{t-1})|\mathcal{E}(x_t),\mathcal{E}(v))\) for all \(t \in [0,1]\) in this latent space. This enables us to recover \(\mathcal{E}(x)\) from random noise given \(\mathcal{E}(v)\).

This part is divided into two phases: training and inference. During the training phase, we concatenate the VAE-encoded thermal latent and RGB image along the channel dimension, then randomly sample at timestep $t$ and train the Unet to predict the noise. After training the model, in the inference phase, we replace the thermal image with standard Gaussian noise and gradually denoise to obtain the predicted thermal image.

\begin{algorithm}[!tbp]
	\caption{Optimization process of LaPIG.} 
	\renewcommand{\algorithmicrequire}{\textbf{STEP1:}}
	\renewcommand{\algorithmicensure}{\textbf{STEP2:}}
	\begin{algorithmic}[1]
		\REQUIRE Training Visible Image Synthesis Pipeline.
		\STATE \textbf{While} training:
		\STATE \hspace{0.3cm} Prepare: ``Face + Face with poses'' \{$\boldsymbol{I}_f, \boldsymbol{I}_{pose}$\}; caption templates \textbf{ct}; ArcFace \textbf{Arc}; Stable Diffusion pipeline \textbf{SD}; LongClip model \textbf{LongClip};
		\STATE \hspace{0.3cm} Forward:  $\{\boldsymbol{I}_{poses}'\}=\textbf{SD}(\textbf{Arc}(I_f), \textbf{LongClip}(\textbf{ct}))$;
		\STATE \hspace{0.3cm} Minimize:  $\tilde{\mathcal{L}}$ formulated in Eq. (\ref{pose});
    \STATE \hspace{0.3cm} Update: LongClip and trainable Unet;
    \STATE \textbf{End While}.

    \ENSURE Training V2T LDM (UNet denoted as $\boldsymbol{\epsilon}_{\boldsymbol{\vartheta}}$).
    \STATE Define: $\alpha_1$, $\alpha_2$, ..., $\alpha_T$ (derived from $\beta_1$, $\beta_2$, ..., $\beta_T$);
		\STATE \textbf{While} training:
    \STATE \hspace{0.3cm} Prepare: ``RGB+Thermal'' \{$\boldsymbol{I}_v, \boldsymbol{I}_t$\};
    \STATE \hspace{0.3cm} Forward:  $\textbf{c}, \boldsymbol{z}_{0} = \text{VAE-Encoder} (\{\boldsymbol{I}_v, \boldsymbol{I}_t\})$; 
    \STATE \hspace{0.3cm} Sample: $t \sim \text{Uniform}(\{1, ..., T\}) $, $\boldsymbol{\epsilon} \sim \mathcal{N}(\mathbf{0}, \mathbf{I})$;
    \STATE \hspace{0.3cm} Calculate: $\boldsymbol{z}_t = \sqrt{\overline{\alpha}_t }\boldsymbol{z}_0 + \sqrt{1-\overline{\alpha}_t } \boldsymbol{\epsilon}$; 
    \STATE \hspace{0.3cm} Minimize: $\mathbb{E}_{\boldsymbol{z}_0, \boldsymbol{\epsilon}, t}\left[\|\boldsymbol{\epsilon}-\boldsymbol{\epsilon}_{\boldsymbol{\vartheta}}\left(\boldsymbol{z}_t, t, \boldsymbol{c}\right)\|_2^2\right]$;
    \STATE \hspace{0.3cm} Update: UNet $\boldsymbol{\epsilon}_{\boldsymbol{\vartheta}}$;
    \STATE \textbf{End While}.
	\end{algorithmic}
	\label{optimization}
\end{algorithm}

\subsection{Training Process}   
During the training phase, we need to proceed in two steps. First, we train the Visible Image Synthesis Pipeline component to generate facial images of different poses while maintaining the identity features of the person. Subsequently, we train the V2T LDM to convert the generated facial images of various poses into corresponding infrared images. Algorithm \ref{optimization} illustrates the optimization process.

Because our dataset includes images of specific individuals in various poses, our Visible Image Synthesis Pipeline can enhance the ability to generate faces in different poses by fine-tuning the LongClip and the corresponding Stable Diffusion Unet, which are two sub-modules. To make poses generated turn out to be specific appearance, we propose a simple and effective strategy that uses the average structural similarity index (SSIM)\cite{1284395} between images as a means to fine-grained control structure similarity:
\begin{equation}
    \tilde{\mathcal{L}}=-\sum_{i=1}^NSSIM(\textbf{Pose}_i, \tilde{\textbf{Pose}}_i),
    \label{pose}
\end{equation}
where $\textbf{Pose}_i$ and $\tilde{\textbf{Pose}}_i$ are reference pose images chosen from our datasets and the generated face pose. We assign 4 angles for faces generated so here $N$ in the equation is 4.

For the part of converting visible images to thermal images, our proposed V2T LDM structure undergoes supervised model training based on the dataset, and the loss function used by the model is as follows:
\begin{equation}
    \mathcal{L}_{V2T} = \mathbb{E}_{\boldsymbol{z}_0, \boldsymbol{\epsilon}, t}\left[\|\boldsymbol{\epsilon}-\boldsymbol{\epsilon}_{\boldsymbol{\vartheta}}\left(\boldsymbol{z}_t, t, \boldsymbol{c}\right)\|_2^2\right]
\end{equation}
where $z_0$ is the concatenated tensor from $\mathcal{E}(x)$ and $\mathcal{E}(v)$.

%% file: sec/4_experiments.tex
\section{Experiments}
In this section, we present the experimental results for each stage of LaPIG. We first introduce the dataset used for evaluation, along with the evaluation metrics and training details, followed by a comprehensive quantitative analysis and qualitative analysis against existing methods.

\begin{table*}[t]
\resizebox{1.0\linewidth}{!}{
\begin{tabular}{ccccccccc}
\toprule
&  \multicolumn{4}{c}{SpeakingFaces\cite{s21103465}}              & \multicolumn{4}{c}{ARL-VTF\cite{poster2021large}}             \\
\cmidrule(lr){2-5}  \cmidrule(lr){6-9}
Variants &  VR@FAR=$0.1$\%($\uparrow$)  &  VR@FAR=$1$\%($\uparrow$)  & SSIM($\uparrow$)   & FID($\downarrow$)   & VR@FAR=$0.1$\%($\uparrow$)  & VR@FAR=$1$\%($\uparrow$)  & SSIM($\uparrow$)   & FID($\downarrow$)     \\
\midrule                                   
\multirow{1}{*}{IP-Adapter\cite{ye2023ip}} & 25.22 & 44.59 & 0.5137 & 50.97 & 20.36  & 36.61   & 0.5277  & 63.49       \\ 
\cmidrule(lr){1-1} \cmidrule(lr){2-5} \cmidrule(lr){6-9}
\multirow{1}{*}{Photomaker\cite{li2024photomaker}} & 29.65 & 52.01 & 0.6058 & 47.22 & 26.42  & 44.11   & 0.5774  & 57.19       \\ 
\cmidrule(lr){1-1} \cmidrule(lr){2-5} \cmidrule(lr){6-9}
\multirow{1}{*}{Arc2face\cite{papantoniou2024arc2face}} & 30.07 & 51.11 & 0.6407 & 43.96 & 24.54  & 47.04   & 0.6288  & 54.66    \\ 
\midrule
\rowcolor[gray]{0.9}
\multirow{1}{*}{\textbf{LaPIG (Ours)}} & \textbf{34.33} & \textbf{53.21} & \textbf{0.6648} & \textbf{42.78} & \textbf{27.66}  & \textbf{49.32}   & \textbf{0.6343}  & \textbf{50.77}    \\ 
\bottomrule
\end{tabular}
}
\caption{{\bf Comparison with other identity-preserving generation methods. } ($\uparrow$) indicates that higher values are better, and ($\downarrow$) indicates that lower values are better. The best results are highlighted in bold. The results in the table pertain to four metrics across two datasets. Each variant represents the use of a specific model to replace the corresponding functional module in our entire pipeline.}
\label{tab:ablation}
\end{table*}

\begin{table*}[t]
\resizebox{1.0\linewidth}{!}{
\begin{tabular}{ccccccccc}
\toprule
&  \multicolumn{4}{c}{SpeakingFaces\cite{s21103465}}              & \multicolumn{4}{c}{ARL-VTF\cite{poster2021large}}             \\
\cmidrule(lr){2-5}  \cmidrule(lr){6-9}
Variants &  VR@FAR=$0.1$\%($\uparrow$)  &  VR@FAR=$1$\%($\uparrow$)  & SSIM($\uparrow$)   & FID($\downarrow$)   & VR@FAR=$0.1$\%($\uparrow$)  & VR@FAR=$1$\%($\uparrow$)  & SSIM($\uparrow$)   & FID($\downarrow$)     \\
\midrule                       
\multirow{1}{*}{pix2pix\cite{isola2017image}} &0.60  & 4.17 & 0.2144 & 102.56 & 0.33  & 5.09   & 0.2313  & 122.77     \\
\cmidrule(lr){1-1} \cmidrule(lr){2-5} \cmidrule(lr){6-9}
\multirow{1}{*}{SAGAN\cite{di2019sagan}} & 0.50 &7.66  & 0.2782 & 105.93 & 0.33  & 5.42   & 0.3416  & 111.45 \\
\cmidrule(lr){1-1} \cmidrule(lr){2-5} \cmidrule(lr){6-9}
\multirow{1}{*}{GANVFS\cite{zhang2019ganvfs}} & 5.13 & 16.11 & 0.3481 & 97.28 & 2.63  & 12.97   & 0.3501  & 92.17       \\
\cmidrule(lr){1-1} \cmidrule(lr){2-5} \cmidrule(lr){6-9}              
\multirow{1}{*}{HiFaceGAN\cite{yang2020hifacegan}} & 15.55 & 39.01 & 0.4084 & 60.77 & 20.89  & 41.33   & 0.4752  & 67.88       \\ 
\cmidrule(lr){1-1} \cmidrule(lr){2-5} \cmidrule(lr){6-9}
\multirow{1}{*}{AxialGAN\cite{immidisetti2021axial-gan}} & 19.19 & 43.38 & 0.4418 & 65.19 & 18.62  & 42.86   & 0.4622  & 65.71       \\ 
\cmidrule(lr){1-1} \cmidrule(lr){2-5} \cmidrule(lr){6-9}
\multirow{1}{*}{DDPM\cite{nair2023t2v}} & 22.15 & 47.92 & 0.5164 & 50.38 & 19.87  & 43.51   & 0.5314  & 49.72    \\ 
\midrule
\rowcolor[gray]{0.9}
\multirow{1}{*}{\textbf{LaPIG (Ours)}} & \textbf{34.33} & \textbf{53.21} & \textbf{0.6648} & \textbf{42.78} & \textbf{27.66}  & \textbf{49.32}   & \textbf{0.6343}  & \textbf{50.77}    \\ 
\bottomrule
\end{tabular}
}
\caption{\textbf{Comparison to other prevailing methods on V2T translation.} ($\uparrow$)  indicates that higher values are better, and ($\downarrow$) indicates that lower values are better. The best results are highlighted in bold. The results in the table pertain to four metrics across two datasets. Each variant represents the use of a specific model to replace the corresponding functional module in our entire pipeline.
}
\label{tab:main_results}
\end{table*}

\subsection{Datasets and Evaluation Metrics}
We primarily assess the performance of models using two datasets and four evaluation metrics, which are detailed below.

\textbf{SpeakingFaces Dataset \cite{s21103465} }includes high-resolution thermal and visual images of faces synchronized with audio recordings of subjects speaking approximately 100 phrases, captured from nine angles. For our experiments, we only use the paired thermal and visual images of faces. In total, there are 100 identities for training and 42 identities for testing, resulting in 5,400 data pairs for training and 2,268 data pairs for testing. SpeakingFaces dataset offers a vast, publicly accessible multimodal corpus suitable for machine learning studies that leverage thermal, visual, and auditory streams. The dataset is split into three parts: train set, validation set, and test set, with each set containing unique subjects. The synchronized samples and the wide diversity of subjects render the SpeakingFaces dataset a robust and challenging resource for research endeavors.

\textbf{ARL-VTF Dataset \cite{poster2021large} }contains facial images captured in Long-Wave Infrared and includes settings for aligning faces. The dataset also provides image capture settings to align the faces. To address the severe overexposure in the visible images, we applied a correction method similar to \cite{nair2023t2v}. For our experiments, we created a subset of the dataset with 100 identities showing different expressions for training and 40 identities for testing, resulting in 3,200 training pairs and 985 testing pairs.

\textbf{Evaluation Metric }
To assess our method's performance, we use two evaluation schemes \cite{duan2020cross,mei2022VPGAN}. We focus on the quality of the reconstruction, measured by four metrics. FID \cite{heusel2017fid}, SSIM\cite{1284395}. We also test face verification, comparing our method with others using Rank-1 accuracy and Verification Rates (VR) at 1\% and 0.1\% False Acceptance Rates (FAR). Understanding VR@FAR=1\% and VR@FAR=0.1\% is crucial for evaluating our LaPIG method. These metrics gauge the accuracy of a biometric system, with FAR indicating the likelihood of accepting an incorrect match. Setting FAR at these levels tests the system's precision at specific error rates. All verification tests use the ArcFace system \cite{deng2019arcface}, which is pre-trained for this purpose.



\subsection{Training Details}
We configured the base learning rate of the VQ-VAE at \(4.5 \times 10^{-6}\) for its training, with a downsampling factor of 8 and 500 training epochs. The encoder is used to handle both thermal and visible images. For the diffusion models, we set a base learning rate of \(2 \times 10^{-6}\) and conducted training over 300 epochs. These experiments were carried out on a single NVIDIA RTX 3090. The training process lasted 10 hours with the VQ-VAE and took 5 hours for the diffusion model training. The inference process takes 70 seconds.

\subsection{Quantitative Analysis}
Quantitative analysis is divided into two main parts: a comparison with other identity-preserving generation methods and a comparison with other image translation methods. When evaluating identity-preserving generation performance, we keep the visible-to-thermal model constant, using our proposed V2T-LDM module. Conversely, when comparing image translation methods, we maintain the identity-preserving generation model constant, using our proposed ID preserving generative model.

\textbf{Comparison with other identity-preserving generation methods}. To verify the generation performance of our mode (particularly the ID consistency before and after image generation, as well as performance on the downstream tasks), we compared it with other state-of-the-art (SOTA) methods, including IP-Adapter \cite{ye2023ip}, Photomaker \cite{li2024photomaker}, and Arc2Face \cite{papantoniou2024arc2face}. The FID and SSIM scores were calculated by comparing the corresponding poses of each individual in the original dataset. All methods were evaluated using this approach.

As shown in Table \ref{tab:ablation}, we tested different schemes on the SpeakingFaces dataset and ARL-VTF dataset. It is evident that our model has successfully alleviated the issue of identity inconsistency and image translation cases prevalent in existing face generation and thermal-to-visible face translation methods. This success is attributed to the design of proposed ID preserving generative model and V2T translation module, which enhanced the identity information and translation ability of the whole pipeline. The utilization of each module has proven effective.

\textbf{Comparison with other image translation methods}. To provide a more quantitative evaluation of the methods, we use the metric VR@FAR=1\% as previously mentioned. The quantitative outcomes are presented in Table \ref{tab:main_results}. The VR@FAR metric provides a measure of the system's accuracy at these FAR levels. A higher VR@FAR value indicates a lower number of incorrectly matched identities at the given FAR, thus signifying better performance. It can be seen that we outperform other variants in all metrics, especially the two VR@FAR indicators, where our results are significantly better than those of other methods. These downstream task-related metrics fully demonstrate the versatility of the data generated by LaPIG.

These experimental results not only highlight the superior performance of LaPIG in terms of quantitative metrics but also underscore the realistic nature of the generated visible images. This increase in accuracy is crucial for applications where the correct identification of individuals is paramount, such as in security and surveillance systems. 

\begin{figure}[t!]
\centering
\includegraphics[width=0.47\textwidth]{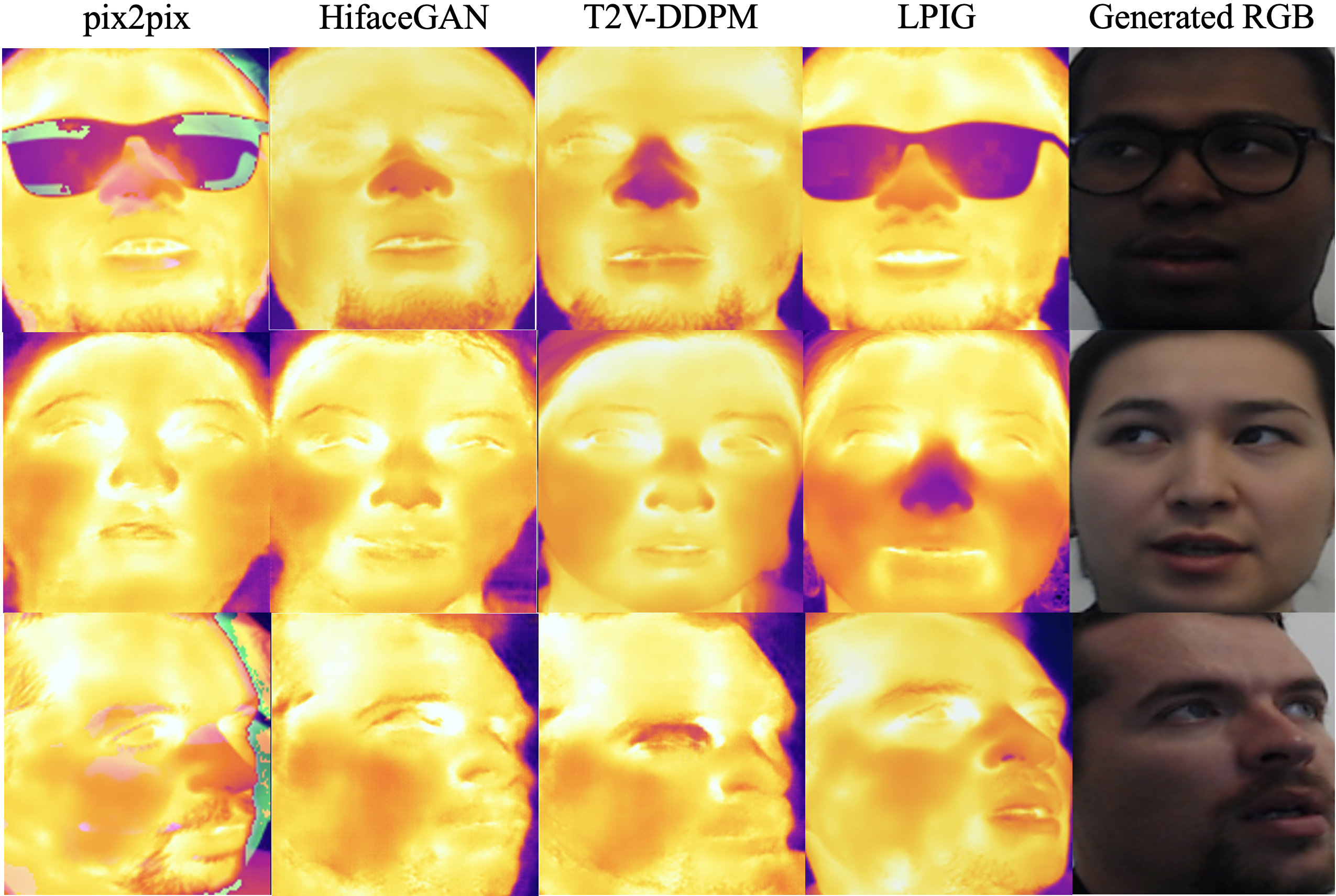}
\caption{{\bf Comparison of translated results. } Each column represents the effect of different methods, with the last column being the generation effect of our proposed identity preserving synthesis model.
}
\label{fig:comparison}
\end{figure}

\begin{figure*}
    \centering
    \begin{subfigure}[b]{0.201\textwidth}
        \centering
        \includegraphics[width=\linewidth]{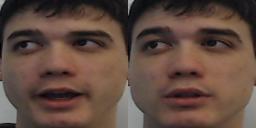}
    \end{subfigure}
    \hfill
    \begin{subfigure}[b]{0.78\textwidth}
        \centering
        \includegraphics[width=\linewidth]{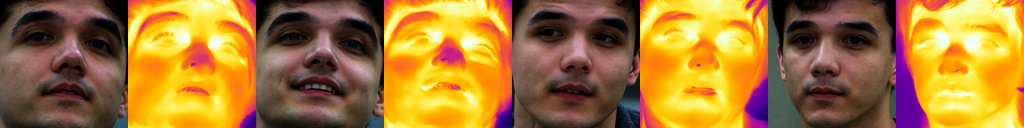}
    \end{subfigure}
    
    \begin{subfigure}[b]{0.201\textwidth}
        \centering
        \includegraphics[width=\linewidth]{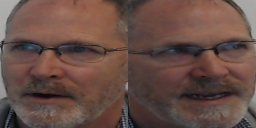}
    \end{subfigure}
    \hfill
    \begin{subfigure}[b]{0.78\textwidth}
        \centering
        \includegraphics[width=\linewidth]{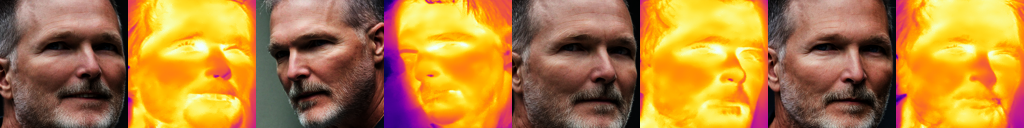}
    \end{subfigure}
    
    \begin{subfigure}[b]{0.201\textwidth}
        \centering
        \includegraphics[width=\linewidth]{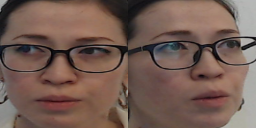}
    \end{subfigure}
    \hfill
    \begin{subfigure}[b]{0.78\textwidth}
        \centering
        \includegraphics[width=\linewidth]{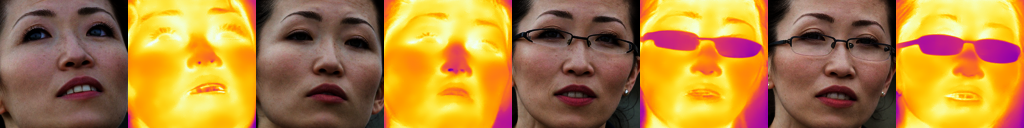}
    \end{subfigure}

    \begin{subfigure}[b]{0.201\textwidth}
        \centering
        \includegraphics[width=\linewidth]{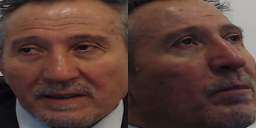}
    \end{subfigure}
    \hfill
    \begin{subfigure}[b]{0.78\textwidth}
        \centering
        \includegraphics[width=\linewidth]{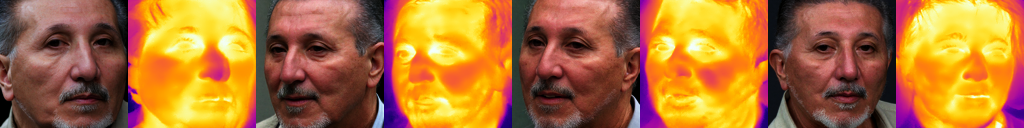}
    \end{subfigure}

    \begin{subfigure}[b]{0.201\textwidth}
        \centering
        \includegraphics[width=\linewidth]{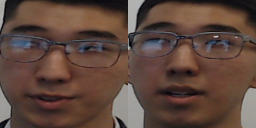}
        \caption{Reference ID Images}
    \end{subfigure}
    \hfill
    \begin{subfigure}[b]{0.78\textwidth}
        \centering
        \includegraphics[width=\linewidth]{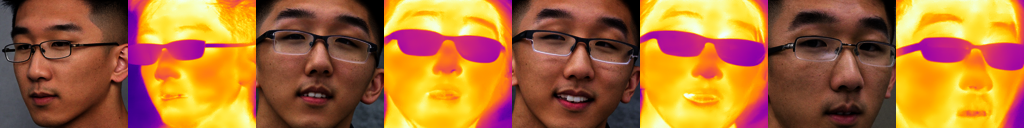}
        \caption{Synthesized Paired Images using LaPIG}
    \end{subfigure}
    
    \caption{\textbf{Examples of Synthesized Paired Data:} The left two images are the input ID images, and on the right is the synthesized paired data by LaPIG.}
    \label{fig:}
\end{figure*}

\subsection{Qualitative Analysis}

Qualitative analysis primarily uses visual comparison to analyze the optimization effects of our various modules and to highlight the advantages over other methods.

As shown in Figure \ref{fig:comparison}, existing face translation methods have their flaws to varying degrees. For instance, they may fail to restore details properly, resulting in blurry facial contours, introduce unwanted eyeglass obstructions, or not properly reflect the presence of glasses where they should be, and sometimes even produce strange background colors. In contrast, our proposed LaPIG has better mitigated these issues, generating faces with richer details, higher clarity, and more accurate facial feature restoration.

\begin{figure}[t!]
\centering
\includegraphics[width=0.47\textwidth]{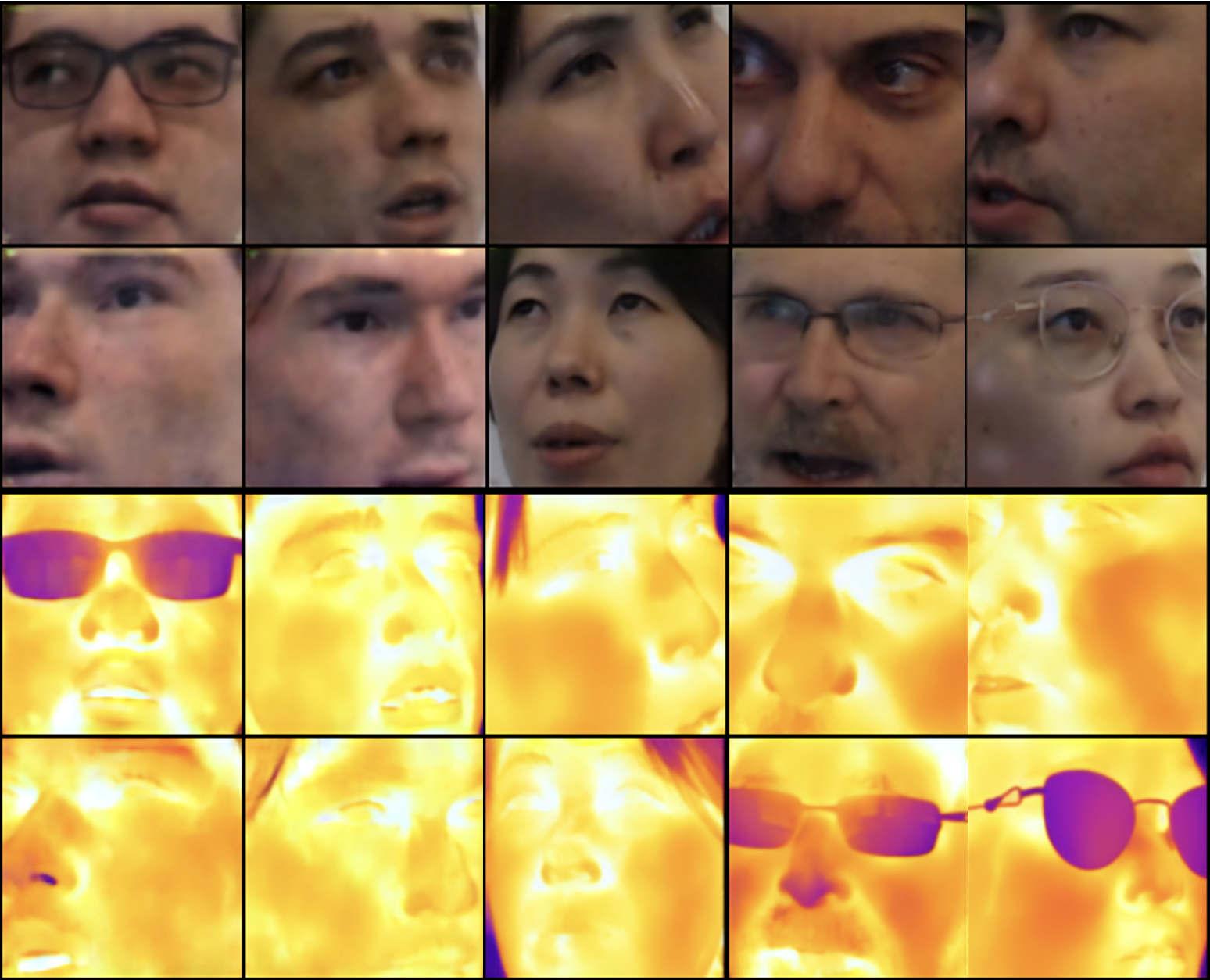}
\caption{{\bf Examples of translated results. } The last two rows present corresponding thermal images produced using the V2T LDM, translated from the visible images presented in the first two rows.}
\label{fig:paired_data}
\end{figure}

More results are shown in Figure \ref{fig:paired_data}, our proposed LaPIG can generate facial images with various poses from multiple angles and obtain the corresponding thermal images, even for different expressions without distortion or deformation. In addition, the model handles occlusions very well. Although glasses appear transparent in the RGB image and are not very distinctive features, the thermal images involving the glasses have all been successfully transformed.

We present an example of synthesized pair images in Figure \ref{fig:}. On the left side, we display the input reference ID images, while the corresponding synthesized paired images are shown on the right. Visually, the synthesized visible images preserve identity-specific features from the reference images, such as the eyebrows, nose, and beard. The translated thermal images maintain high quality as well—the trained translation network accurately captures mouth shapes and positions glasses correctly, demonstrating superior performance compared to existing methods.

In contrast to other methods, our approach delivers high-quality, accurate results that closely match the identity details of real faces. This not only proves the effectiveness of LaPIG but also underscores its robust generalization for paired image generation across varying scenarios. Moreover, LaPIG uniquely facilitates the recognition of generated faces as ``real” within verification systems, further demonstrating its practical superiority for applications demanding high fidelity and identity preservation. To our knowledge, LaPIG represents the first instance of leveraging LDMs for high-fidelity and identity-consistent thermal facial image reconstruction from visible inputs, marking a significant advancement in the field.

%% file: sec/5_conclusion.tex
\section{Conclusion}
In this paper, we introduce LaPIG, an innovative framework designed to generate high-quality paired visible-thermal images with the assistance of LLMs. Our method comprises three stages: first, we propose an identity-preserving synthesis model, enhanced by integrating a Long-CLIP encoder and the LLaVA assistant to generate textual prompts, ensuring identity retention in the synthesized images. Next, we utilize an LDM-based translation network to transform visible images into thermal images. Finally, we generate captions with the LLaVA assistant, enabling multiview and diverse paired image synthesis. Extensive experiments demonstrate the effectiveness of our approach.

LaPIG can also be applied to efficiently construct a high-quality paired visible-thermal dataset, reducing both time and financial costs compared to manual collection. The synthesized paired data can further support downstream tasks like facial recognition. For example, a recognition system trained on this data could recognize faces in visible images during the day and in thermal images at night, enhancing security under various lighting conditions. Future work could explore expanding LaPIG’s applications in real-world scenarios.